\documentclass[twocolumn, color]{interact}
\usepackage{url}
\usepackage{multicol}
\usepackage{makecell}
\usepackage{epstopdf}
\usepackage[caption=false]{subfig}
\usepackage[dvipsnames, svgnames, table]{xcolor}
\usepackage[most]{tcolorbox}
\newtcolorbox{shadedabstract}{
    colback=MyTerracotta!30!white, 
    colframe=MyTerracotta!30!white, 
    boxrule=0.5pt, 
    sharp corners, 
    width=\linewidth,
    left=6pt, right=6pt, top=6pt, bottom=6pt,
    boxsep=0pt,
    before upper=\vspace{1pt}\upshape, 
}
\definecolor{MyTerracotta}{RGB}{204, 128, 119}
\usepackage[numbers,sort&compress]{natbib}
\bibpunct[, ]{[}{]}{;}{n}{,}{,}

\renewcommand\bibfont{\fontsize{9}{11}\selectfont}
\usepackage{helvet}

\theoremstyle{plain}

\theoremstyle{definition}

\theoremstyle{remark}

\definecolor{MyTitles}{RGB}{166, 40, 30}
\usepackage{titlesec}
\titleformat{\section}
  {\normalfont\Large\bfseries\color{MyTitles}} 
  {\thesection}                                    
  {0.5em}                                          
  {}                                               
\titleformat{\subsection}
  {\normalfont\large\bfseries\itshape\color{MyTitles}} 
  {\thesubsection}                                 
  {0.5em}
  {}
\titleformat{\subsubsection}
  {\normalfont\normalfont\itshape\color{MyTitles}} 
  {\thesubsection}                                 
  {0.5em}
  {}
\titlespacing*{\subsubsection}{0pt}{3.25ex plus 1ex minus .2ex} {0pt}

\begin{document}


\title{Time-to-Injury Forecasting in Elite Female Football: A DeepHit Survival Approach}

\author{
\name{Victoria Catterall\textsuperscript{a}, Cise Midoglu\textsuperscript{b}, Stephen Lynch\textsuperscript{a},\thanks{CONTACT S. Lynch Email: s.lynch@lboro.ac.uk}}
\affil{\textsuperscript{a}Loughborough University, Epinal Way, Loughborough, UK; \textsuperscript{b}Forzasys AS, Oslo, Norway}
}

\maketitle

\begin{shadedabstract}
\noindent{\Large\bfseries\textcolor{MyTitles}{Abstract}}

Injury occurrence in football poses significant challenges for athletes and teams, carrying personal, competitive, and financial consequences. While machine learning has been applied to injury prediction before, existing approaches often rely on static pre-season data and binary outcomes, limiting their real-world utility. This study investigates the feasibility of using a DeepHit neural network to forecast time-to-injury from longitudinal athlete monitoring data, while providing interpretable predictions. The analysis utilised the publicly available SoccerMon dataset, containing two seasons of training, match, and wellness records from elite female footballers. Data was pre-processed through cleaning, feature engineering, and the application of three imputation strategies. Baseline models (Random Forest, XGBoost, Logistic Regression) were optimised via grid search for benchmarking, while the DeepHit model, implemented with a multilayer perceptron backbone, was evaluated using chronological and leave-one-player-out (LOPO) validation. DeepHit achieved a concordance index of 0.762, outperforming baseline models and delivering individualised, time-varying risk estimates. Shapley Additive Explanations (SHAP) identified clinically relevant predictors consistent with established risk factors, enhancing interpretability. Overall, this study provides a novel proof of concept: survival modelling with DeepHit shows strong potential to advance injury forecasting in football, offering accurate, explainable, and actionable insights for injury prevention across competitive levels.
\end{shadedabstract}


\begin{multicols}{2}
\section*{Introduction}

Football is considered the most popular sport globally. It exerts a profound social, cultural, and financial influence on communities, and there are approximately 265 million active players worldwide (Hulteen et al., 2017).  Yet injury remains a universal challenge in the sport, costing professional clubs an estimated £45 million per season in medical care and lost revenue (Eliakim et al., 2020). For players, injuries bring pain, reduced motivation, and anxiety over recovery and career prospects (Borg, Falzon \& Muscat, 2021) whilst fan engagement, and team performance and cohesion also suffer (Hägglund et al., 2013). Protecting athletes from injury is therefore both a moral and strategic responsibility for clubs and coaches. There is a substantial body of research addressing injury prevention in football; however, less than 25\% of this literature focuses specifically on female players (John et al., 2025). This significant gender disparity in research focus necessitates immediate attention to ensure evidence-based injury mitigation strategies are equally available for female athletes.

Machine learning (ML)-based injury forecasting has emerged as an effective tool in elite sport, surpassing traditional statistical methods (Rossi, Pappalardo \& Cintia, 2021). Effective ML implementation, however, requires large-scale, high-quality datasets to meaningfully capture complex, multimodal relationships necessary for real-time, individualized athlete forecasts (Van Eetvelde et al., 2021; Ghosh et al., 2023). High-quality data collection is a significant challenge in resource-limited settings such as women’s and youth football (Angelova-Igova \& Naydenova, 2023). Even where comprehensive monitoring datasets exist in elite football, competitive and financial constraints typically restrict the open sharing of anonymised data, thereby limiting knowledge dissemination (Eliakim et al., 2020). The advancement of ML-driven forecasting critically relies on increasing the availability of open-access datasets. To maximise utility and model generalisability, curators must prioritise well-labelled features, minimised data missingness (via high adherence), objective injury reporting, and diverse player demographics (e.g., sex, age, competition level) (Mandorino et al., 2023). Predictive accuracy is enhanced by including longitudinal internal and external load measures (e.g., RPE, heart rate) (Zhu, Jiang \& Yamamoto, 2022), supplemented by subjective metrics like readiness or soreness (Castilla-López \& Romero-Franco, 2024).

Due to the rarity of large-scale athlete monitoring datasets available, most models still rely on static pre-season data to predict injury outcome in the subsequent season. These models are rarely validated in female cohorts and lack clinical utility as they are unable to adjust to mid-season environmental changes (Leckey et al., 2024). Binary classification algorithms like Random Forest and XGBoost often yield the best performance (Leckey et al., 2024), however these models cannot capture temporal dependencies or cumulative workload effects central to injury development (Delecroix et al., 2018). Time-series modelling offers greater clinical utility by identifying short-term risk fluctuations that allow timely training adjustments rather than season-wide caution. For example, survival analysis models estimate the likelihood of injury within specific time windows while accounting for censored observations (Turkson, Ayiah-Mensah \& Nimoh, 2021). Although, previous Cox regression survival models in sport have only modelled time to first-injury-event (Mahmood, Ullah \& Finch, 2014), so they require further development to handle recurrent injuries and finer temporal granularity (Jarmann, 2023). The DeepHit survival model (Lee, Yoon \& Schaar, 2020) has been shown to outperform traditional survival models in the medical sector. It has been shown to make accurate short-term prognoses for cancer thrombosis, and cardiovascular disease outcomes (Shen et al., 2024; Mantha et al., 2024; Wang et al., 2024), so warrants further development in the context of sports injury too.

The application of DeepHit modelling could be enhanced by improving model explainability to increase trust amongst players and practitioners (Rossi, Pappalardo \& Cintia, 2021). Without explainability, coaches cannot identify the causes of increased injury risk or adjust training accordingly. Because injury risk factors are highly individual and fluctuate weekly (Mandorino et al., 2023), unguided interpretation can lead to misinformed decisions. Even accurate models lose practical value if their predictions cannot be meaningfully applied in real time - and may even heighten anxiety or hesitant play when players are labelled “high risk” which increases injury likelihood (Slimani et al., 2018). Ethical risks also arise from potential bias, which could unjustifiably exclude players from training or competition (Musat et al., 2024). 

Therefore, the primary aim of this study is to establish the feasibility and predictive capability of a DeepHit survival model for sport-injury forecasting using longitudinal athlete monitoring data. It promotes the use of the SoccerMon dataset (Midoglu et al., 2024) as a rich, female-centric resource, featuring both objective and subjective metrics, to address the gender disparity in sports medicine literature. This work delivers an accurate and interpretable ML model designed to directly inform proactive injury risk management protocols and optimise athlete wellbeing.

\section*{Methodology}

\subsection*{Data Collection}
Data was sourced from the Team B files within the public SoccerMon dataset (Midoglu et al., 2024). Additional data access was granted to Team B's objective injury reports which were used instead of the self-reported injury in the public dataset to improve predictive accuracy and minimise noise. Athlete data was collected via two systems. Subjective data was self-reported by players using the PmSys Athlete Monitoring System (Johansen et al., 2020), whilst objective training data was collected via the STATSports APEX GNSS tracking system, using GPS systems and heart rate monitors to gather metrics on speed, positioning, and exertion. The dataset, collected in 2020–2021 from two elite Norwegian women’s first-division football teams, contains training, match, and injury data (see Table 1). It has been anonymised with informed consent obtained from all participants (Midoglu et al., 2024). 

\begin{table*}
\centering
\caption{Features within the SoccerMon dataset used in model construction.}
\label{tab:model-features}
\resizebox{\textwidth}{!}{
\footnotesize 
\begin{tabular}{p{3cm} p{7cm} p{5.5cm}}
\toprule
\textbf{Feature} & \textbf{Definition} & \textbf{Notes} \\
\midrule
Daily load & Daily training load & Sum of session RPE per day \\
\addlinespace
ATL & Acute training load & Average training load over past 7 days \\
\addlinespace
Weekly load & Weekly training load & Sum of training load over past 7 days\\
\addlinespace
Monotony & Training load monotony & Day to day variability in training load\\
\addlinespace
Strain & Training strain & Training load x monotony\\
\addlinespace
ACWR & Acute:chronic workload ratio & Measure of consistency between short- and long-term training load\\
\addlinespace
CTL28 & 28 day chronic training load & Sum of training load over past 28 days\\
\addlinespace
CTL42 & 42 day chronic training load & Sum of training load over past 42 days\\
\addlinespace
Fatigue & Self-reported fatigue level & \\
\addlinespace
Mood & Self-reported mood & \\
\addlinespace
Readiness & Self-reported readiness to train & \\
\addlinespace
Sleep duration & Self-reported sleep duration & Supposed to be measured by the athlete objectively\\
\addlinespace
Soreness & Self-reported muscle soreness & \\
\addlinespace
Stress & Self-reported stress level & \\
\addlinespace
RPE & Rate of perceived exertion & Subjective measure of how challenging a training session was \\
\addlinespace
sRPE & Session RPE & RPE x session duration\\
\addlinespace
duration\_subj & Subjective duration & How long the athlete judged the session to be\\
\addlinespace
duration\_obj & Objective duration & How long the session lasted\\
\addlinespace
Speed\_km\_h\_mean & Average running speed (kilometers per hour) & GPS measurement\\
\addlinespace
Speed\_km\_h\_max & Maximum running speed (kilometers per hour) & GPS measurement\\
\addlinespace
Speed\_km\_h\_std & Standard deviation in running speed & Quantifies within-session speed variability\\
\addlinespace
sp\_lir\_p & Proportion of training spent at low intensity running speed & \\
\addlinespace
sp\_lir\_t & Time spent at low intensity running speed & \\
\addlinespace
sp\_lir\_d & Distance covered at low intensity running speed & \\
\addlinespace
sp\_mir\_p & Proportion of training spent at medium intensity running speed & \\
\addlinespace
sp\_mir\_t & Time spent at medium intensity running speed & \\
\addlinespace
sp\_mir\_d & Distance covered at medium intensity running speed & \\
\addlinespace
sp\_hir\_p & Proportion of training spent at high intensity running speed & \\
\addlinespace
sp\_hir\_t & Time spent at high intensity running speed & \\
\addlinespace
sp\_hir\_d & Distance covered at high intensity running speed & \\
\addlinespace
sp\_spr\_p & Proportion of training spent sprinting & \\
\addlinespace
sp\_spr\_t & Time spent sprinting & \\
\addlinespace
sp\_spr\_d & Distance covered whilst sprinting & \\
\addlinespace
Distance & Distance covered overall (km) & GPS measurement\\
\addlinespace
distance\_per\_min & Average distance covered per minute & distance \/ duration\_obj\\
\addlinespace
subjective\_missingness\_7d & Proportion of subjective questionnaires that the athlete did not complete in the past 7 days & Calculated feature - not in original dataset\\
\addlinespace
past\_injury\_count & Cumulative count of injuries present in the dataset per athlete & Calculated feature - not in original dataset\\
\bottomrule
\end{tabular}}
\end{table*}

\subsection*{Data Preparation}
Individual feature files were consolidated into a single dataset for analysis. Exploratory data analysis was performed using histograms to examine feature distributions and identify outliers arising from GPS or heart rate monitor signalling errors. Maximum speeds of over 32 kilometers per hour were excluded as physiologically implausible (Vescovi, 2018). Sessions longer than 200 minutes were excluded as signalling errors (Silva et al., 2023), as were running distances exceeding 16 kilometres. In a typical match, players cover between 9 and 11 kilometres (Datson et al., 2017); however, values up to 16 kilometres were retained because histogram inspection indicated a substantial number of valid observations within this range. 

Previous injuries increase the risk of future injury, so a cumulative count of previous injuries was generated (Gashi et al., 2023). A second derived feature was a weekly indicator capturing the proportion of missing responses in players’ daily subjective questionnaires. It was hypothesised that incomplete reporting may signal disengagement from training or recovery protocols, which has been shown to increase injury risk (Matijašević and Rajković Vuletić, 2024). After feature derivation, missing values were imputed using three different trial methods. The first method replaced missing entries with each player’s median value for that feature. The second was a bespoke approach that positioned a player’s values relative to their teammates over the preceding two weeks and generated replacements that maintained this relative standing on days with missing data. The third applied linear interpolation imputation. Histograms and kernel density estimates (KDEs) were produced to compare feature distributions before and after imputation, and correlation coefficients assessed the preservation of associations with injury occurrence. Features that continued to exhibit substantial missingness following imputation were removed to reduce dimensionality in the machine learning models (Salam et al., 2021).

\subsection*{Baseline Models}
To confirm the strength of the dataset for predictive injury-risk forecasting, three common high-performing baseline models (XGBoost, Random Forest and Logistic Regression) were built and trained to use current-day data to predict next-day injury-risk (Leckey et al,. 2024). They were optimised using a grid search method and evaluated using a bespoke weighted formula which prioritised the models' F1 score and recall but also considered precision and AUC (Figure 1). Accuracy was not emphasized due to the highly imbalanced dataset used. Both greedy feature selection and recursive feature elimination were tested to optimise the features used in the models to maximise the efficiency, generalisability, and accuracy of the prediction models by reducing dimensionality (Camattari et al., 2024; Theerthagiri and Vidya, 2022). Principal component analysis was also considered but rejected since it reduces model interpretability and can negatively affect the performance of non-linear ML models (Reddy et al., 2020). To prevent data leakage, models were trained on datasets split chronologically (Yang, Li and Jiang, 2024), and oversampling of the injured class was applied to training data to address class imbalance (Mujahid et al., 2024). 

\begin{figure*}
    \centering
    \includegraphics[width=0.85\linewidth]{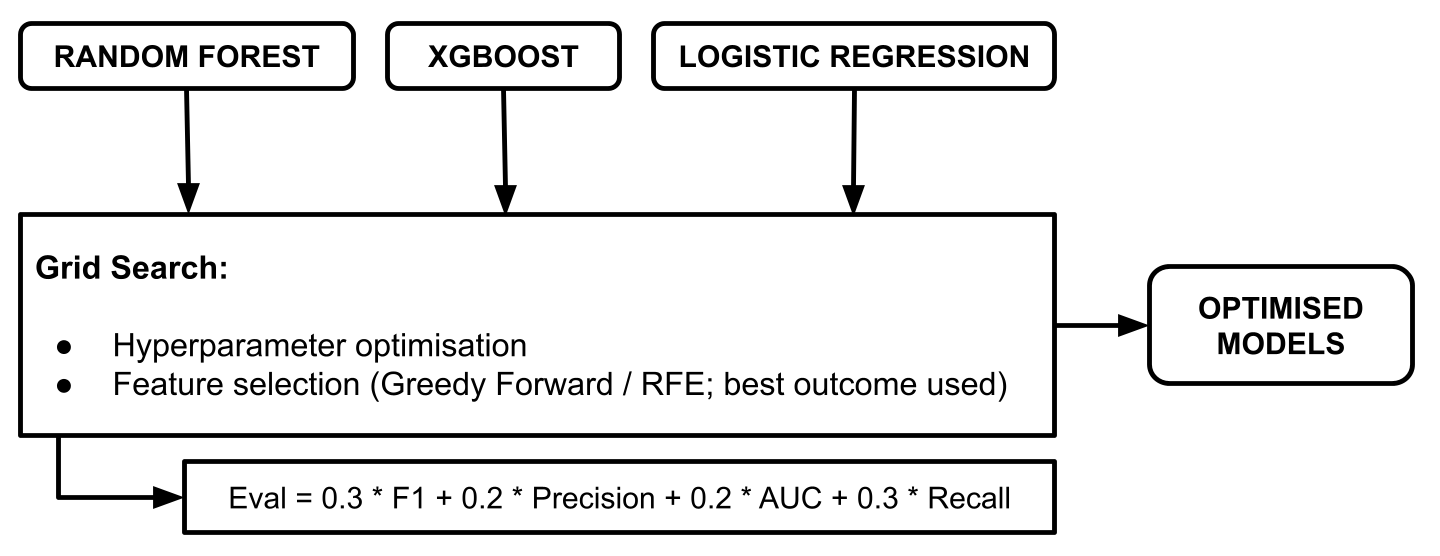}
    \caption{Grid Search method used, including custom evaluation formula and different feature selection techniques.}
    \label{Figure 1}
\end{figure*}

\subsection*{Baseline Look-Back and Forecasting Windows}
Once the initial baseline models were created, the forecasting horizon window was iteratively expanded to 3, 5, 7, 10, and 14 days in length, as was look-back data input window. For each window, rolling averages of the features were computed to capture recent trends, and outcome labels were assigned a value of 1 if an injury occurred within the subsequent prediction horizon. These were also evaluated using F1 score, AUC, precision and recall and the same grid search method was used to optimise the model architecture. 

\subsection*{DeepHit Model}
A standard DeepHit model was initialised with a multilayer perceptron (MLP) backbone rather than a recurrent neural network (RNN) backbone due to data limitations. The dataset was reformatted to include a time-to-event column, where the value decreased from 7 (seven days before injury) to 1 (one day before injury). An event indicator was also created, where 1 denoted an injury and 0 indicated censoring (i.e., no injury observed during the data collection period). The model was implemented with a 21-day input window and a 7-day prediction horizon, consistent with a design that had previously demonstrated success in survival modelling on this dataset (Jarmann, 2023). All features were standardised using a standard scaler to ensure comparability across variables and to improve neural network training stability.

The DeepHit model was first evaluated using a version trained on a chronological split of the dataset, replicating how the system would be applied in practice. This evaluation was then repeated for each dataset generated by each imputation method. The concordance index (C-index) was selected as the primary evaluation metric here, as it measures the model’s ability to correctly rank players by relative injury risk - an outcome directly relevant to coaching decisions (Park et al., 2021) and is the standard metric in survival analysis evaluation (Longato, Vettoretti, and Di Camillo, 2020). To further assess generalisability, a leave-one-player-out (LOPO) validation procedure was implemented; the model was iteratively trained on all but one player and then tested on the excluded individual, with the process repeated for every player. This method is particularly valuable in smaller, player-level datasets, as it reduces the likelihood of the model overfitting to individual injury histories and provides a stronger indication of its ability to adapt to unseen players (Lin et al., 2023). Performance per player was assessed using the C-index, and predicted survival times were converted into daily risk estimates, which were compared against observed injury events using player-level plots. 

\subsection*{Explainability}
To investigate differences in model performance across players following LOPO validation, a correlation analysis was conducted to identify any underlying systematic patterns. For players with the highest C-index values, SHAP analyses were performed to determine their key individual injury risk factors both across the full season and on specific high-risk days. These insights can inform the development of player profiles to support individualized long-term training strategies and guide coaches in making targeted, short-term adjustments to training protocols on days of elevated injury risk.

\section*{Results}

\subsection*{Exploratory Data Analysis}
Exploratory data analysis found patterns of missingness by player and feature, and the distribution of injuries within the dataset. The dataset comprised 37 players tracked over two seasons, with adherence to data collection varying across individuals. During this period, 43 injuries were recorded; 24 injuries were classified as acute and 19 as overuse, with the thighs and knees representing the most frequently injured body parts. After preparation, the dataset comprised 39 features with varying degrees of missingness across players and variables. More injuries were recorded in the second season than in the first, and only 21 players were present from the start of data collection. There was more missingness in subjective than objective features and 15 out of the 37 players had approximately half their data missing on training days (Figure 2). 

\begin{figure*}
    \centering
    \includegraphics[width=\linewidth]{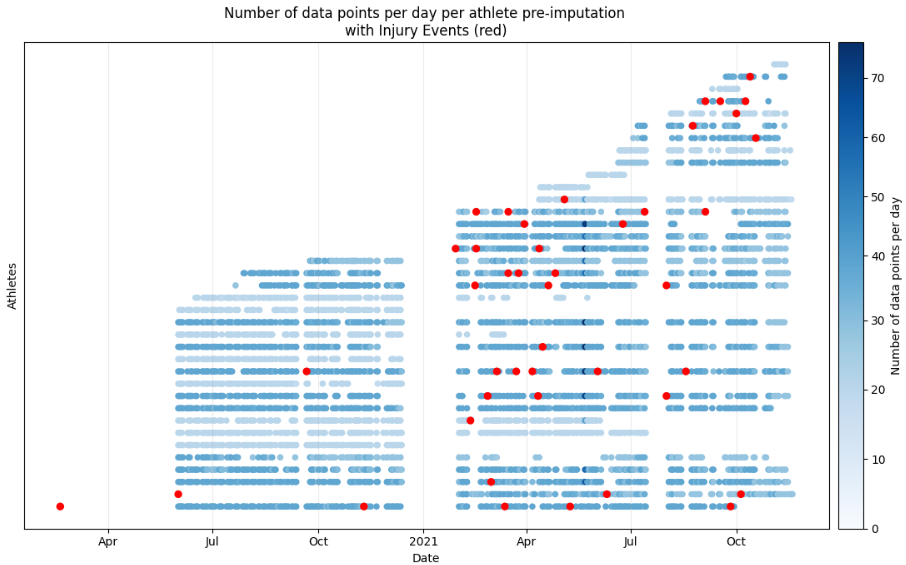}
    \caption{Distribution of data points in the cleaned dataset across 37 players over two football seasons (322 days recorded). Red circles indicate injury occurrences (n = 43).}
    \label{Figure 2}
\end{figure*}

\subsection*{Imputation Methods}
The dataset comprised 4,449 player-date observations. Three imputation strategies were evaluated to address data missingness: median imputation, a bespoke formula, and linear interpolation. Median imputation resolved the highest proportion of missing values but substantially altered the overall data distribution. The bespoke formula imputed the fewest missing values, but did not significantly disturb the data distribution or alter univariate correlations with injury occurrence. Linear interpolation preserved the dataset distribution most closely and was the second most effective at reducing missingness, also without materially affecting univariate correlations with injury occurrence. Used in combination with the DeepHit model, the bespoke formula resulted in the best predictive performance.

\subsection*{Baseline Models}
After grid search optimisation, the Random Forest model achieved the highest predictive performance when using current day data to predict next-day injury risk (F1=0.533, AUC=0.779, precision=1.000, recall=0.364). XGBoost followed behind (F1=0.429, AUC=0.876, precision=1.000, recall=0.271), and Logistic Regression demonstrated the worst predictive performance (F1=0.071, AUC=0.758, precision=0.037, recall=0.833), see Table 2. Model optimisation involved iterating through various input and output window combinations. While models offering the greatest predictive lead time and the shortest input window are highly advantageous in a practical sports setting as they allow for long-term planning even for new players, models with a forecasting window larger than one day led to significant reductions in performance, so this line of investigation was not pursued further.

\begin{table*}
\centering
\caption{Performance metrics for XGBoost, Random Forest, and Logistic Regression models using previous-day data to predict injury occurrence.}
\label{tab:model-performance}
\footnotesize 
\begin{tabular}{l c c c c c c c >{\raggedright\arraybackslash}p{5cm}}
\toprule
\textbf{Model} & \textbf{F1 score} & \textbf{Precision} & \textbf{Recall} & \textbf{AUC} & \textbf{No. Features} & \textbf{\thead{Look-back \\ Window}} & \textbf{\thead{Forecasting \\ Window}} \\
\midrule
Random Forest & 0.533 & 1.000 & 0.364 & 0.779 & 19 & 1 day & 1 day\\
\addlinespace
XGBoost & 0.429 & 1.000 & 0.273 & 0.876 & 18 & 1 day & 1 day \\
\addlinespace
Logistic Regression & 0.071 & 0.037 & 0.833 & 0.758 & 8 & 1 day & 1 day\\
\bottomrule
\end{tabular}
\end{table*}

\subsection*{DeepHit Model}

\subsubsection*{Chronological Train/Test Split}
The model was trained on the first 80 percent of the dataset in chronological order. Training data imputed with linear interpolation achieved a C-index of 0.660 which was increased to 0.762 with training data imputed using the bespoke method. 

\subsubsection*{LOPO Split}
With LOPO training, the interquartile range of C-indices across players was 0.192 (Figure 3), indicating a wide variability in model generalisability. The highest predictive performance observed produced a C-index of 0.974, while for others, the model performed poorly. Correlation analyses revealed a weak positive correlation between C-index and the number of sessions tracked (r = 0.44). No correlation was observed between C-index and the number of injuries (r =–0.08). Higher within-player variance in training load and behavioural patterns reduced predictive performance, as did players who were clear outliers relative to their teammates. 

\begin{figure*}
    \centering
    \includegraphics[width=\linewidth]{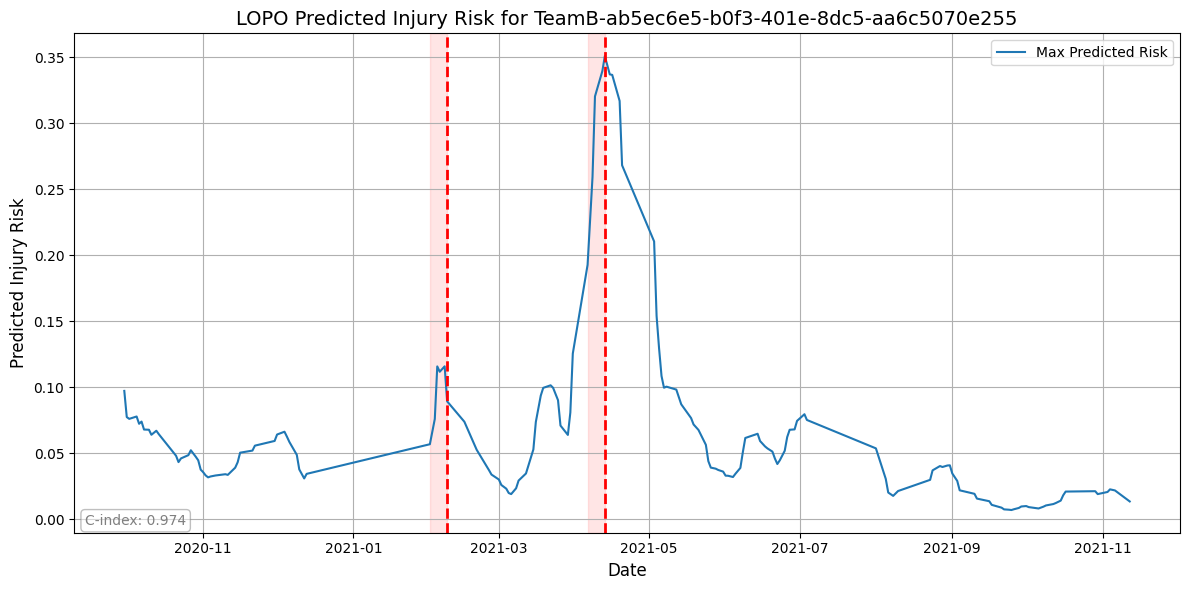}
    \caption{Predicted injury risk over time by a DeepHit model trained using a LOPO method. Red dotted lines indicate injury occurrences.}
    \label{Figure 3}
\end{figure*}

\subsubsection*{Model Explainability - Case Study}
As a proof of concept, the player with the highest C-index after LOPO training was analysed. The risk predicted by the model successfully peaked ahead of a real injury occurrence twice. Across the data-collection period, this player's key injury risk factors were high stress, high high-intensity running volume, low mood, and low sleep quality. Similar features were involved on a specific day flagged as high risk, but with high fatigue also serving as a primary contributor to the high-risk prediction (Figure 4). Notably, this player sustained an acute thigh injury following this period of elevated predicted risk. 

\begin{figure*}
    \centering
    \includegraphics[width=0.85\linewidth]{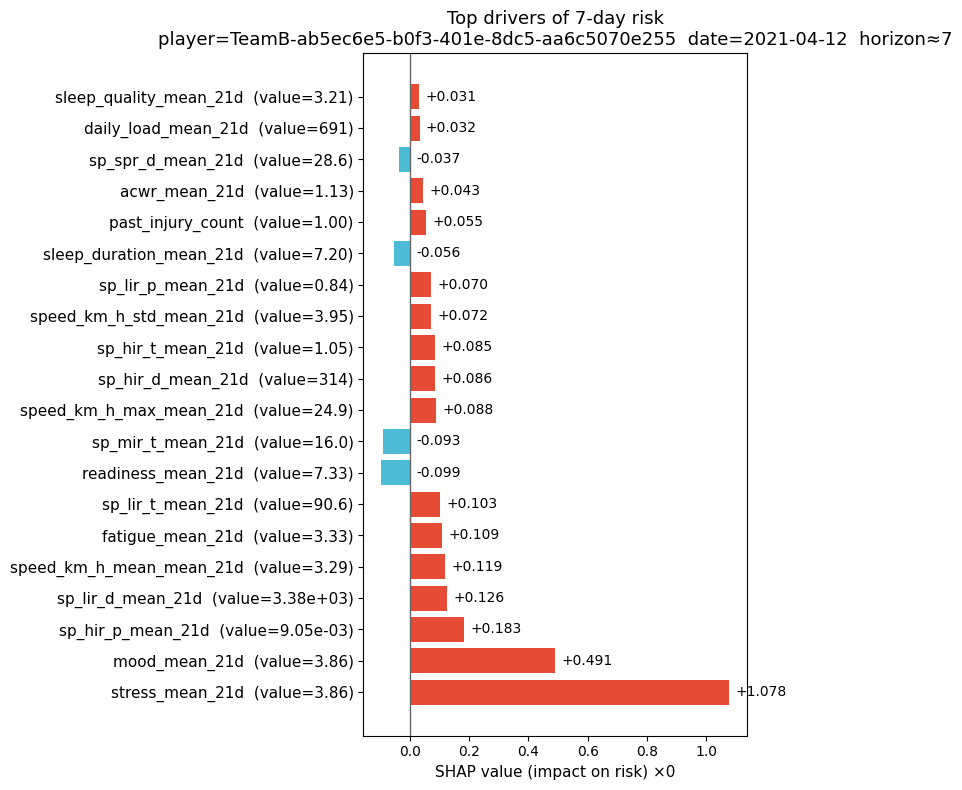}
    \caption{Results of SHAP analysis determining the features that most influenced a prediction of elevated injury risk for one player on one day.}
    \label{Figure 4}
\end{figure*}

\section*{Discussion}
This study offers novel insight into the application of a DeepHit model for sport-injury forecasting with findings that outperform baseline models in accuracy and clinical utility. 

\subsection*{Forecasting Insights}
\subsubsection*{Baseline Models}
Out of three baseline models trained chronologically, the Random Forest model performed the best. It produced no false positives while correctly identifying 36.4\% of injury cases, and was the only model to achieve an F1 score above 0.5 (0.533). Although an F1 score of 0.533 would typically be considered modest in broader ML contexts (Wingrove et al., 2020) - within the constraints of this highly imbalanced and complex dataset, it represents a promising outcome. Whilst the XGBoost model also performed well, the superior performance of the Random Forest may reflect its ensemble architecture being better suited to managing nonlinear relationships and class imbalance than Logistic Regression (linear) and XGBoost (boosting-based) architectures.

These findings align with previous research identifying Random Forest and XGBoost as leading models for sports injury prediction (Leckey et al., 2024). Their performance is often similar but varies by sport and predictive task. For instance, XGBoost outperformed in forecasting next-season injuries in ice hockey (Luu et al., 2020), while Random Forest performed best in a comparable baseball study (Karnuta et al., 2020). Such variation likely reflects sport-specific injury mechanisms and contexts, such as differing contact levels. The average AUC of sports injury models is 0.69 (Leckey et al., 2024), which both models surpassed here when using current-day data to predict next-day injuries - indicating that the dataset captured meaningful injury-risk patterns. However, the F1 score of 0.533 was below the literature average of 0.73 (Leckey et al., 2024), suggesting a need to improve recall and overall predictive balance. 

\subsubsection*{DeepHit Model}
The best DeepHit model iteration used data imputed via the bespoke formula and achieved a concordance index (C-index) of 0.762. This indicates strong discriminative ability, with nearly 80\% of player-day pairs correctly ranked by relative injury risk. Within survival analysis literature, C-index values above 0.7 are considered acceptable, while values above 0.8 are regarded as strong (Longato, Vettoretti \& Di Camillo, 2020). Thus, achieving 0.762 is a highly encouraging outcome given the challenging nature of the dataset and the relatively limited degree of model tuning undertaken. Although little research has examined the influence of imputation strategies on DeepHit specifically, it is plausible that the bespoke approach outperformed linear interpolation because it preserved sharp fluctuations in the data that often precede injuries. 

The DeepHit model was trained and evaluated using a LOPO approach. Whilst this setup is less representative of real-world deployment, it offers valuable insight into how well the model generalises to unseen players - a key consideration when integrating new squad members. A C-index was calculated for each held-out player, revealing substantial variability (IQR = 0.192), and indicating room for improvement in generalisation. Performance ranged widely (up to C-index = 0.974), suggesting the model captured risk dynamics that were highly predictive for some players, particularly those with consistent reporting and clear links between training load and injury outcomes. But the same model struggled with players whose data were irregular, sparse, or distinct from team norms. These individuals may represent outliers due to unique training characteristics or reporting styles. Incorporating contextual factors such as playing position, role-specific workloads, or player archetypes could reduce this variability (Shimakawa et al., 2025), whilst personalisation strategies, such as fine-tuning with individual data, may further enhance robustness across diverse player profiles. 

\subsection*{Explainability}
The deployment of ML models in real-life sports medicine practice requires the development of trust from practitioners and the output of actionable coaching insights. It is therefore essential that models are designed to deliver explanations alongside their injury-risk predictions in terms of what key features are influencing risk so that coaches can alter training protocols accordingly. 

\subsubsection*{Feature Importance}
One method of enhancing explainability is via feature importance analysis. With the SoccerMon dataset, the most influential features aligned with previous injury-risk findings (Ekstrand et al., 2023), including average running speed, soreness, monotony, and prior injury count, supporting the model’s validity. However, several results diverged from expectations. For instance, time [of training session] was more important than expected, while acute to chronic workload ratio (ACWR), sprint-related metrics, and subjective stress contributed little. These discrepancies warrant cautious interpretation. The prominence of time may reflect a proxy for session type if higher-risk sessions occur at specific times of day, whilst the limited impact of sprint metrics could stem from low sprint thresholds that underrepresented meaningful sprint events. ACWR effects may also have been diluted by the 21-day input window. Notably, subjective missingness emerged as a key predictor, possibly indicating that players withhold responses when anticipating injury or that low engagement with monitoring reflects reduced adherence to training or recovery protocols - both potential risk factors. Collectively, these findings highlight the importance of thoughtful feature design and interpretation in injury prediction models and demonstrate SHAP’s value in revealing non-intuitive risk-factor relationships. 

\subsubsection*{Explainability of The DeepHit Model}
Using the DeepHit model and SHAP analysis, feature importance was ranked for a high-risk prediction day following LOPO training. Elevated stress, high-intensity running, low mood, and increased fatigue raised predicted injury-risk, while greater readiness and longer sleep had modest protective effects. This pattern preceded an acute thigh injury and aligns with prior research linking training load, fatigue, and psychological stress to acute injuries (Sampson et al., 2019). Furthermore, the key features on this day differed from those identified globally for the same player, suggesting the model can detect context-specific risk fluctuations rather than relying solely on stable, season-long patterns. This represents one of the first applications of survival modelling in sport to offer explainability at the individual player-day level, providing a proof of concept for personalised, daily injury-risk interpretation. Such insights could help practitioners make targeted adjustments to training, for example, addressing elevated stress as a short-term risk signal, or soreness as a marker of cumulative fatigue. 

\subsection*{Limitations and Future Work}
\subsubsection*{Experimental Design}
The experimental design is strong in its use of varied imputation methods, chronological test-train split, and evaluation methods (Altalhan, Algarni and Alouane, 2025). Future work should expand on this foundation by replicating real-world deployment of ML models by gradually feeding the model data, and evaluating the updated predictions over time (rolling-origin evaluation). Further, hyperparameter tuning and feature engineering should be investigated too, with additional contextual factors such as playing position being taken into account.

\subsubsection*{DeepHit Model}
The DeepHit model demonstrated considerable promise as a proof of concept for time-to-injury prediction with encouraging predictive performance and strong concordance with existing literature on injury risk factors. Its implementation here is based on a multilayer perceptron (MLP) backbone but it is thought that Dynamic-DeepHit models with a recurrent neural network (RNN) backbone could provide superior predictive performance and better handling of temporal data. Future work should evaluate this hypothesis, but would require rich data with evenly spaced temporal inputs and a very low degree of missingness. The missingness and irregular temporal spacing between entries in the SoccerMon dataset made this infeasible here, so the curation and dissemination of high-quality datasets is essential for the progression of this research avenue.

Future investigations would also benefit from developing the model to inherently handle missing data; accounting for non-training days and irregular gaps; developing per-player risk thresholds; incorporating time since last injury to better model reinjury risk; and extending the prediction horizon beyond one week. Coupled with a suitable interface for coaches, these steps would help move the model from a research prototype towards a deployable decision-support tool.

\subsubsection*{SoccerMon Dataset}
The dataset itself represents a major strength of this project. It provides multimodal information combining objective load and physiological data with subjective well-being reports, and is based on female professional athletes: an underrepresented group in literature (Paul et al., 2022). Furthermore, because the data sources were relatively low-cost and easily collectable, this approach could be replicated by resource-limited teams at lower playing levels, expanding its potential impact beyond elite sport. Therefore, future studies would benefit from reproducing the features present in the SoccerMon dataset. They should also focus on maximising athlete adherence  within their investigations, minimising the number of incomplete subjective reports would ensure even coverage between players and increase predictive performance. Additional features, such as menstrual cycle information, biomarkers, or routine movement screening data, could further enrich the dataset and improve model performance, though these may be harder to collect consistently. Overall, expanding datasets to include more injuries, players, teams, and demographics would both improve the robustness of the DeepHit model and broaden its applicability.

\section*{Conclusion}
This study demonstrates the feasibility of forecasting sports injuries using athlete monitoring data and a DeepHit machine learning model. The SoccerMon athlete monitoring dataset is used and evaluated, which is based on elite female footballers and contains objective and subjective features (Midoglu et al., 2024). Median imputation, linear interpolation, and a bespoke formula are trialed to address data missingness, with the bespoke imputation formula resulting in the best DeepHit performance. Feature importance and SHAP explainability are used to investigate the features influencing injury-risk prediction both on a season-wide and a day-to-day scale. 

The DeepHit survival model, benchmarked against Random Forest, XGBoost, and Logistic Regression on the SoccerMon dataset, achieved the strongest performance with a concordance index of 0.762 using the bespoke imputation method - accurately ranking injury risk in nearly 80\% of cases. Unlike baseline classifiers, DeepHit generated individualised, time-varying risk estimates, explained by SHAP analysis, that aligned with known injury mechanisms such as load, stress, and recovery. It offers both predictive power and interpretability, which would be critical for practitioner trust and adoption.

Future research should focus on expanding the accessibility of larger, multi-team, longitudinal athlete monitoring datasets, with SoccerMon setting a good example of the feature set required for strong predictive performance. Minimising missingness and providing evenly spaced temporal data should be prioritised in order to construct and evaluate advanced model architectures (e.g. Dynamic DeepHit) with enhanced personalisation strategies to enhance generalisation. Real-time deployment and practitioner-facing interfaces will also be key to translation into practice.

Overall, this work highlights the potential of survival-based deep learning to advance sports injury forecasting. With such promising predictive performance and detailed model explainability, this work could serve as the foundation for further development and broader validation. In future, DeepHit modelling has the capacity to play a significant role in supporting athlete health, optimising training load management, and addressing the persistent challenge of sports injuries within both elite and grassroots settings.

\end{multicols}

\section*{Acknowledgement(s)}

Victoria Catterall would like to thank Professor Stephen Lynch for his help and encouragement throughout the development of this paper, and Dr Cise Midoglu and her team in Norway for their insight and collaboration on this project.

\section*{Disclosure Statement}

The authors declare no conflict of interest.

\section*{Funding}

This research recieved no external funding.

\section*{Notes on Contributors}

Conceptualisation: VC, 
Methodology: VC \& CM, 
Software: VC, 
Validation: SL, 
Formal analysis: VC, 
Investigation: VC \& CM, 
Data curation: CM \& VC, 
Writing - original draft: VC, 
Writing - review \& editing: CM \& SL \& VC, 
Visualisation: VC, 
Supervision: SL

\section*{Data Availability Statement}
SoccerMon data is openly available in a public repository that issues datasets with DOIs (\url{https://zenodo.org/records/10033832}).
Objective injury report data is not openly available due to legal restrictions.
Code is openly available in a public repository that does not issue DOIs (\url{https://github.com/simulamet-host/soccermon-deephit}).

\end{document}